\documentclass{article}

\usepackage{geometry}
\usepackage{graphicx} 
\usepackage{caption, subcaption}
\usepackage{amsmath,amsthm,amssymb,amsfonts}
\usepackage{array}
\usepackage{url}
\usepackage{pdfpages}
\usepackage{threeparttable}
\usepackage[backend=biber,style=numeric]{biblatex}
\title{Particle swarm optimization with Applications to Maximum Likelihood Estimation and Penalized Negative Binomial Regression}
\author{Junhyung Park, Sisi Shao, Weng Kee Wong}

\author{Junhyung Park$^{1,+,*}$, Sisi Shao$^{+,2}$, Weng Kee Wong$^{2}$ 
$^{1}$ \\United States Naval Academy, Annapolis, MD\\
$^{2}$Department of Biostatistics, University of California, Los Angeles, CA 90095, U.S.A\\
$^+$These authors contribute to the paper equally.}
 
\begin{document}

\maketitle
\date{}
\bigskip
\begin{abstract}
General purpose optimization routines such as nlminb, optim (R) or nlmixed (SAS) are frequently used to estimate model parameters in nonstandard distributions. This paper presents Particle Swarm Optimization (PSO), as an alternative to many of the current algorithms used in statistics. We find that PSO can not only reproduce the same results as the above routines, it can also produce results that are more optimal or when others cannot converge.  In the latter case, it can also identify the source of the problem or problems.  We highlight advantages of using PSO using four examples, where: (1) some parameters in a generalized distribution are unidentified using PSO when it is not   apparent or computationally manifested using routines in R or SAS; (2) PSO can produce estimation results for the log-binomial regressions when current routines may not;  (3) PSO provides flexibility in the link function for binomial regression with LASSO penalty, which is unsupported by standard packages like GLM and GENMOD in Stata and SAS, respectively, and (4) PSO provides superior MLE estimates for an EE-IW distribution compared with those from the traditional statistical methods that rely on moments.  
\end{abstract}

\noindent%

\section{An Overview of PSO}
\hspace{0.5cm} Metaheuristics, and in particular, nature-inspired metaheuristic algorithms,  is increasingly used across disciplines to tackle challenging optimization problems \cite{cui2024applications}.  They may be broadly categorized  swarm based or evolutionary based algorithms.
 Some examples of the former are particle swarm optimization and  competitive swarm optimizer (CSO) and examples of the latter are genetic algorithm (GA) and the differential evolution. 
The statistical community is probably most aware of GA and simulated annealing (SA) but they are many others that have recently proven more popular  in engineering and computer science. \cite{kumar2023} provides the latest comprehensive review of nature-inspired metaheuristic algorithms.

 Particle Swarm Optimization (PSO) is a stochastic numerical search algorithm that has garnered widespread acclaim among scholars across various disciplines. Initially conceptualized by Kennedy \& Eberhart in 1995 \cite{kennedy}, PSO draws inspiration from the biological swarm behavior of birds, emulating their leaderless yet highly coordinated approach to locate food targets. This intriguing aspect of animal swarms is meticulously captured in PSO.  \cite{marini} provides an instructive tutorial on PSO and its applications that include  chemometrics, signal alignment and robust PCA projection pursuit. Recently, applications of PSO for solving various statistical problems include variable selection \cite{cui2024applications}, bioinformatics \cite{cui2022single,cui2022d, ermis2023estimation}, design \cite{wong2015modified, chen2015minimax}, and artificial intelligence in medicine \cite{shiai}. Accordingly, our overview of PSO is brief as we focus on the aims of the paper, which highlighted in the abstract.

PSO  is a general purpose optimization tool that requires virtually no technical assumptions for it to work quite efficiently.  For a given function $f$ defined on a user-selected compact set $S$, PSO tackles the optimization problem by finding $x_{max}\in S$ such that
$f(\mathbf{x}_{max})\geq f(\mathbf{x}) ~~~\forall \mathbf{x} \in S$.  The space $S$ can be high-dimensional with or without several linear or nonlinear constraints.

 PSO works by generating an initial swarm of particles of a given size to search for the optimum by communicating with one another  and varying their positions as they seek to converge to an optimum. At each iteration, each particle flies to its next position by flying with a velocity and the movement of the $i^{th}$ particle is governed by two equations:
\begin{equation}\mathbf{x}_{i}^{t+1} = \mathbf{x}_{i}^{t} + \mathbf{v}_{i}^{t+1}     \label{eq:blah}
\end{equation}

\begin{align}& \mathbf{v}_{i}^{t+1} = \phi^{t}\mathbf{v}_{i}^{t}+c_{1}u_{1}^{t} \left( \mathbf{p}_{i}^{t}-\mathbf{x}_{i}^{t}\right)+ c_{2}u_{2}^{t}\left(\mathbf{g}^{t}-\mathbf{x}_{i}^{t}\right).\label{eq:velo} 
\end{align}

In the above equations, $\phi,c_{1},c_{2}\in\mathbb{R}^{1},$\text{and} $u_{1}^{t}$ \text{and} $u_{2}^{t}$ are independent uniform variates from the the interval $U[0,1]$.The superscript $t$ is the iteration number and the velocity of a particle at time $t$ for particle $i$ is denoted by $\mathbf{v}_{i}^{t}$. Adding the velocity to the position of the particle $\mathbf{x}$, yields a new position (\ref{eq:blah}). Further, $\mathbf{p}_{i}^{t}$ is the best position visited by particle $i$ through the $t^{th}$ iteration and is called the \textit{Personal best}, i.e, $\mathbf{p}_{i}^{t} = max_{x} \left\{f(\mathbf{x}_{i}^{1}),f(\mathbf{x}_{i}^{2}), \ldots,f(\mathbf{x}_{i}^{t})\right\}$ for particle $i=\{1\ldots N\}$. The notation $\mathbf{g}^{t}$ is the best of all such $\mathbf{p}_{i}$'s through the $t^{th}$ iteration, hence, $\mathbf{g}^{t} = max_{p} \{f(\mathbf{p}_{1}^{1}),f(\mathbf{p}_{2}^{1}),
\ldots,f(\mathbf{p}_{N}^{1}),\ldots,f(\mathbf{p}_{1}^{t}), \allowbreak f(\mathbf{p}_{2}^{t}),
\ldots , f(\mathbf{p}_{N}^{t})\}$. This position found at iteration $t$ is called the \textit{Global best}.  At termination, it is hope that  \textit{Global best} is the solution sought for the optimization problem.  
There have been many modifications and variations of PSO proposed to improve the efficiency and stability of convergence, and optimizing PSO to work well for specifications of swarm applications. Examples include defining a $\mathbf{V}_{max}$ which sets an upper bound to the velocity of a particle in any iteration (\cite{shahzad}, \cite{ghalia}). Varying inertia weights are implemented in \cite{chatterjee} and \cite{yufeng}, and different velocity update rules are seen in \cite{liu} and \cite{poli}. Various search constraints are also illustrated in \cite{pulido} and \cite{pulido2}. There is also much discussion of PSO parameter selection in the literature. We refer the reader to sample references such as \cite{zhang}, \cite{zheng}, \cite{shi98_2} and \cite{campana}.

One of the goals of this paper is to show that even the most basic specification of PSO can be robustly applied to estimation problems in statistics. In particular, we set $c_{1}=c_{2}=2$, and $\chi=1$ as seen in \cite{shi00}. The inertia weight is set to decrease linearly from around one to zero ($\phi^{t}=1-\frac{t}{M}$, $t=\{1\ldots M\}$ ). Variations to these settings, in particular $c_{1}=1$ and $c_{2}=3$, or a non-linear inertia weight ($\phi^{t}=1-\log{t}/\log{M}$, $t=\{1\ldots M\}$ ) did not give practically different results in convergence.

The basic implementation of PSO is as follows:
\begin{itemize}
	\item Specify tuning parameters
	\item Randomly initialize particles' positions over the search space
	\item Initialize velocities
	\item Evaluate the fitness $f(.)$ of each particle and determine $\mathbf{g}^{1}$ and $\mathbf{p}_{i}^{1}$
	\item This gives what is needed to compute $\mathbf{v}_{i}^{2}$
	\item Compute $\mathbf{x}_{i}^{2}$ using (\ref{eq:blah})
	\item Re-evaluate the new fitness of each particle
	\item Update $\mathbf{p}_{i}^{2}=f(\mathbf{x}_{i}^{2})$ if $f(\mathbf{x}_{i}^{2})>f(\mathbf{x}_{i}^{1})$ or otherwise $\mathbf{p}_{i}^{2}=\mathbf{p}_{i}^{1}$
	\item Update $\mathbf{g}^{1}$ if a better global best is found or otherwise $\mathbf{g}^{2}=\mathbf{g}^{1}$
	\item This gives everything needed to compute $\mathbf{v}_{i}^{3}$
	\item Finally, compute $\mathbf{x}_{i}^{3}$ and repeat the above process for a set number of iterations, or until a tolerance criteria is met.
\end{itemize}

This paper initializes the particle velocity as zero, and particle positions are initialized in a uniform distribution between predetermined lower and upper bounds of the search space. When a particle strays outside the search bounds [L,U], it is either (depending on the example) re-randomized uniformly into the interval [L,U], or randomized in a small neighborhood of L or U that lies within the interval [L,U].

In Section \ref{sec2}, we verify that PSO can reproduce solutions to non-trivial estimation problems obtained by using existing methods. In Section \ref{sec3}, coefficients of more complex distributions are estimated. In the process, we find that larger likelihood values are obtained by PSO compared to a Newton-based algorithms used by SAS or R. Here, we also discover that PSO gives us an insight about the objective function that otherwise would have been overlooked while using standard statistical software.  Section \ref{sec4}  shows how PSO can overcome problems with the choice of the initial values and convergence and consequent failures in log binomial regressions. We summarize and conclude in Section \ref{sec5}.

\section{Reproducing Known Results}
\label{sec2}
Our primary goal is to demonstrate that Particle Swarm Optimization (PSO) can consistently replicate results derived from conventional statistical packages. To illustrate this, we have selected an example problem related to the estimation of parameters within a generalized Weibull distribution, as motivated by Bourguignon et al. (2014) \cite{bourg}. The original results were obtained using the SAS procedure \texttt{proc nlmixed}. Additionally, through exercises that are not detailed in this report, we have ascertained that PSO can dependably achieve accurate parameter estimates for simpler statistical challenges. These include Ordinary Least Squares, Logistic Regression, and the estimation of means and covariance matrices for multivariate normal data.

The Weibull distribution $(x>0)$ with respective cdf and pdf
\begin{align*}
	F(x) &=1-\exp\{-\alpha x^{\beta}\} \textnormal{~~and~~} \\
	f(x) &=\alpha\beta x^{\beta-1} \exp \{-\alpha x^{\beta}\},~~\alpha,\beta>0.
\end{align*}
is generalized by replacing $x$ with the odds ratio $[G(x)/\bar{G}(x)]$, where $\bar{G}(x)=1-G(x)$. Here, $G(x;\boldsymbol{\xi})$ is some continuous distribution of choice with density $g(x;\boldsymbol{\xi})$ and vector of parameters $\boldsymbol{\xi}$.
The new cdf is obtained by letting $u=-\alpha t^\beta$,
\begin{align} F(x;\alpha,\beta,\boldsymbol{\xi}) &=\int_{0}^{\frac{G(x;\boldsymbol{\xi})}{\bar{G}(x;\boldsymbol{\xi})}}
	\alpha\beta t^{\beta-1} \exp\{-\alpha t^{\beta}\} dt  \nonumber \\
	&=\int_{0}^{-\alpha\left(\frac{G}{\bar{G}} \right)^\beta} -e^u du
	=1-e^{-\alpha\left(\frac{G}{\bar{G}} \right)^\beta}.  \label{eq:weigcdf}
\end{align}
Letting $k(x)=G(x)/\bar{G}(x)$, the new pdf is

\begin{align}
	&f(x;\alpha,\beta,\boldsymbol{\xi}) = \alpha k'(x) \exp \left\{-\alpha k(x)^{\beta} \right\} \nonumber \\
	&= \alpha\beta g(x;\boldsymbol{\xi})\frac{G(x;\boldsymbol{\xi})^{\beta-1}}{\bar{G}(x;\boldsymbol{\xi})^{\beta+1}} \exp\left\{ -\alpha \left[ \frac{G(x;\boldsymbol{\xi})}{\bar{G}(x;\boldsymbol{\xi})} \right]^{\beta} \right\}.  \label{eq:weigpdf}
\end{align}

Depending on a choice of $G$, the pdf in (\ref{eq:weigpdf}) can yield a variety of distributions\cite{bourg}. Bourguignon et al. (2014) calls this the Weibull-$G$ family of distributions. The motivation behind the generalization is as follows \cite{cooray}. Suppose $Y$ is a lifetime random variable with distribution $G(y;\boldsymbol{\xi})$. Then the odds ratio that an individual with random lifetime $Y$ will die or fail at time $x$ is $G(x;\boldsymbol{\xi})/\bar{G}(x;\boldsymbol{\xi})$. If $X$ represents the variability on this odds ratio, and follows a Weibull model with parameters $\alpha$ and $\beta$, then
\[
Pr\left(X\leq \frac{G(x;\boldsymbol{\xi})}{\bar{G}(x;\boldsymbol{\xi})} \right)
=F(x;\alpha, \beta, \boldsymbol{\xi}).
\]
In other words, we have a Weibull-$G$ distribution when $G$ describes random lifetimes and odd ratios are Weibull distributed.

Based on real data, we find parameter estimates by maximizing the log-likelihood functions from three different distributions: Weibull-Exponential (WE), Exponentiated-Weibull (EW) and Exponentiated-Exponential (EE) where all distributions have $\alpha, \beta, \lambda>0$. The first is obtained by choosing $G(x;\lambda)=1-e^{-\lambda x}$. According to (\ref{eq:weigcdf}), the WE cdf is

\[
F_{WE}(x;\alpha,\beta,\lambda)=1-\exp \left[ -\alpha(e^{\lambda x}-1)^{\beta}  \right], ~~x>0.
\]
We note that if $\beta=1$ and $\alpha=\theta/\lambda$, $(\theta>0)$, the Gompertz distribution obtains \cite{gompertz}.
The corresponding pdf and log-likelihood function are

\begin{align*}
	f_{WE}(x;\alpha, \beta,\lambda) &=\alpha\beta\lambda(1-e^{-\lambda x})^{\beta-1}
	e^{\lambda\beta x -\alpha(e^{\lambda x}-1)^{\beta}}\\
	l(\alpha,\beta,\lambda\mid x_{1}\ldots x_{n})&=n\ln(\alpha)+n\ln(\beta)+n\ln(\lambda)+\sum^{n}_{i=1}(\beta-1)\ln(1-e^{-\lambda x_{i}})  \\
	&+\sum^{n}_{i=1} \left[ \lambda\beta x_{i}-\alpha(e^{\lambda x_{i}}-1)^{\beta} \right].
\end{align*}

 Further, the EW and EE distributions are taken as given in \cite{mudholkar} and \cite{gupta} respectively. Their densities are as follows and log-likelihood functions are obtained in a straight forward manner as WE

\begin{align*}
	f_{EW}(x;\alpha,\beta,\lambda) &=\alpha\beta\lambda^{\beta}x^{\beta-1}
	e^{-(\lambda x)^{\beta}}(1-e^{-(\lambda x)^{\beta}})^{\alpha-1}
	\\
	f_{EE}(x;\alpha,\lambda)&=\alpha\lambda e^{-\lambda x}(1-e^{-\lambda x})^{\alpha-1}.
\end{align*}

\noindent The data represented by ${x_{1}\ldots x_{n}}$ are from Smith and Naylor (1987) \cite{smith}. There are 63 observations of the strengths of 1.5cm glass fibers, originally obtained by workers at the UK National Physical Laboratory.

All WE parameters in PSO were initialized over $U[0,4]$. The lower bound of zero comes by definition, and the upper bound of 4 is informed by the result from \cite{bourg}. Note that a more generous upper bound causes no problem for the final PSO result. All elements have positivity constraints as required by the definition of the distribution, and elements are re-randomized near zero if they stray negative ($U[0,0.5]$). For this particular application, we note that the update equation $\mathbf{x}_{i}^{t+1}\ =\ \mathbf{x}_{i}^{t}\ +\  \mathbf{v}_{i}^{t+1}$ could result in a negative value, meaning a negative velocity. When this happens, we  re-randomize the negative element around zero using a uniform distribution: \textit{U[0,0.5]}.

\begin{table}[t]
	\centering
	\scalebox{1}{
		\begin{tabular}{|c c >{\centering\arraybackslash}p{2cm} c c c >{\centering\arraybackslash}p{2.5cm}|}
			\hline
			Method & Swarm Size & Number of Iterations & $\alpha$ & $\beta$ & $\lambda$ & Log-Likelihood\\
			\hline
			NLMIXED &       &       &   0.014750     &  2.8796 &  1.0178   & -14.4020744   \\
			PSO         & 100  & 200  &   0.014742   &  2.87936 & 1.01793 &  -14.4020744   \\
			PSO         & 50   & 200  &   0.014865   &  2.88085 & 1.01705 &  -14.4020746   \\
			PSO         & 100  & 100  &   0.014879   &  2.88427 & 1.01513 &  -14.4020771   \\
			PSO         & 50   & 150  &   0.014777   &  2.88057 & 1.01723 &  -14.4020745   \\
			\hline
		\end{tabular}
	}
	\caption{Weibull-Exponential: specifications of swarm sizes, number of iterations and results. Row 2 to 5 show that PSO with different swarm sizes and number of iterations produce comparable (or the same) results compared with the NLMIXED function in SAS. Further, Rows 2 and 3 show that a larger swarm size provides a more optimal value. Rows 3 and 5 suggest that with the same swarm size, 150 iterations seems sufficient for PSO to converge in this problem.}
	\label{table:we}
\end{table}

\begin{table*}[t]
	\centering
	\scalebox{0.9}{
		\begin{tabular}{|c c c >{\centering\arraybackslash}p{2cm} c c c >{\centering\arraybackslash}p{2.5cm}|}
			\hline
			Method & Model & Swarm Size &Number of Iterations & $\alpha$ & $\beta$ & $\lambda$ & Log-Likelihood\\
			\hline
			NLMIXED &  EW   &      &       &   0.6712     &  7.2846  &  0.5820  &   -14.6755268    \\
			PSO         &  EW   &1000  & 1500  &   0.671243   &  7.28459 &  0.58203 &   -14.6755222    \\
			NLMIXED &  EE   &      &       &   31.349     &          &  2.6116  &   -31.38347218   \\
			PSO         &  EE   &1000  & 1500  &   31.3489    &          &  2.61157 &   -31.38347214   \\
			\hline
		\end{tabular}
	}
	\caption{Extremely close estimates from PSO and those reported in \cite{bourg} for the EW and EE distributions obtained by NLMIXED.  PSO used a swarm size of 1000 and  1500 iterations.} 
	\label{table:comparison}
 
\end{table*}

The far right column in Table \ref{table:we} shows that the maximized log-likelihood value found by PSO is virtually the same as what is found in SAS. The parameter estimate found by PSO is also the same with SAS up to a decimal precision. The greatest likelihoods are achieved with a swarm size of 1000 at 2000 iterations. This is not to say that a different combination of swarm size and iterations (say, 2000 and 1000 respectively) would yield an equally fit solution.

After testing various combinations of swarm size and iterations, we noticed it is difficult to reliably achieve an optimum log-likelihood value beyond the $100,000^{th}$ decimal precision with a swarm size below 500. In other words, there was a minimum number of particles required to achieve an optimal log-likelihood for a given decimal precision. An arbitrarily high decimal precision can be achieved beyond what is shown in the table by further increasing swarm size and/or iterations.

 the remaining two distributions (EW and EE) presented in Table \ref{table:comparison} show again that PSO matches the SAS results. The PSO particles for the EW parameters and EE parameters are initialized over $U[0,10]$ and $U[0,40]$ respectively, both with positivity constraints and rules just as described for WE above. The choice of these bounds was informed by the results known \textit{a priori} from SAS. We will see in the next section that the initialization range does not necessarily have to contain the solution for PSO.

\section{Discovering Redundant Parameters}
\label{sec3}
Unlike in Section \ref{sec2}, PSO found markedly different results when applied to more complicated, 5-parameter mixture distributions. Two examples are the Weibull-Burr-XII distribution (WBXII) and the Beta-Burr-XII distribution (BBXII).

The first belongs to the Weibull-$G$ family summarized in Section \ref{sec2}, where the parent distribution $G$ in (\ref{eq:weigpdf}) is chosen to be $G(x;s,k,c)=1-[1+(x/s)^{c}]^{-k}$. This is the cdf of the Burr-XII distribution introduced by \cite{zimmer}. According to (\ref{eq:weigcdf}), the cdf of WBXII is
\begin{align*}
	F_{WBXII}(x;\alpha,\beta,s,k,c)=1-\exp \left\{ -\alpha \left[ \left( 1+ \left( \frac{x}{s} \right) ^{c} \right) ^{k}-1 \right] ^{\beta} \right\}.
\end{align*}
As done similarly in Section \ref{sec2}, it is straight forward to show that the log-likelihood is
\begin{align*}
	l(\alpha,\beta,s,k,c &\mid x_{1}\ldots x_{n})= n[\ln(\alpha)+\ln(\beta)+\ln(c)+\ln(k)-c\ln(s)]\\
	&+(c-1)\sum^{n}_{i=1}\ln(x_{i})-(1-k)\sum^{n}_{i=1}\ln \left( 1+ \left( \frac{x_{i}}{s} \right) ^{c} \right)\\
	&  -\alpha\sum^{n}_{i=1}\left[ \left( 1+ \left( \frac{x_{i}}{s} \right) ^{c} \right) ^{k}-1 \right] ^{\beta} +(\beta-1)\sum^{n}_{i=1}\ln \left( \left( 1+ \left( \frac{x_{i}}{s} \right) ^{c} \right) ^{k}-1 \right).
\end{align*}

The second distribution (BBXII) is defined by \cite{paranaiba}. The derivation of BBXII is the same as in WBXII, except that the integrand of the cdf is a Beta distribution pdf (instead of Weibull) and that the upper limit of the integral (so-called the ``baseline function'')  is not an odds ratio. More concretely, \cite{eugene} defined a class of generalized Beta distributions by
\begin{align*}
	 F(x)=I_{G(x)}(\alpha,\beta)=\frac{1}{B(\alpha,\beta)}\int_{0}^{G(x)}w^{\alpha-1}(1-w)^{\beta-1}dw,\enspace \textnormal{where}\enspace\alpha,\beta>0.
\end{align*}

Here $I_{\ast}(\alpha,\beta)=B_{\ast}(\alpha,\beta)/B(\alpha,\beta)$ is the ratio where $B_{\ast}(\alpha,\beta)=\int_{0}^{\ast}w^{\alpha-1}(1-w)^{\beta-1}dw$ denotes the incomplete beta function and $B(\alpha,\beta)=\Gamma(\alpha)\Gamma(\beta)/\Gamma(\alpha+\beta)$ denotes the regular beta function with gamma functions $\Gamma(\cdot)$. The corresponding pdf is
\begin{equation}
	f(x)=\frac{g(x)}{B(\alpha,\beta)}G(x)^{\alpha-1} \{ 1-G(x) \}^{\beta-1}
	\label{eq:bbxiipdf}
\end{equation}
where $g(x)={dG(x)}/{dx}$. To obtain the BBXII distribution, we choose $G$ to be the cdf of a Burr XII distribution
\[
G(x,s,k,c)=1-\left[ 1+ \left(\frac{x}{s}\right)^{c} \right]^{-k}
\]
such that the cdf of the BBXII is
\[
F(x)=\frac{1}{B(\alpha,\beta)}\int_{0}^{1-\left[ 1+ \left(\frac{x}{s}\right)^{c} \right]^{-k}}w^{\alpha-1}(1-w)^{\beta-1}dw
\]
and the log-likelihood function for a sample of size  ${x_{1}, x_{2}\ldots x_{n}}$ is obtained similarly to WBXII:
\begin{align*}
	l(\alpha,\beta,s,k,c &\mid x_{1}\ldots x_{n})= n[\ln(c)+\ln(k)-c\ln(s)]+(c-1)\sum^{n}_{i=1}\ln(x_{i}) \\
	&-n\ln(B(\alpha,\beta))-(k\beta+1)\sum^{n}_{i=1}\ln \left( 1+ \left(\frac{x_{i}}{s}\right)^{c}\right)\\
	&+(\alpha-1)\sum^{n}_{i=1} \ln \left( 1-\left[ 1+ \left(\frac{x_{i}}{s}\right)^{c} \right]^{-k}  \right).
\end{align*}

We use PSO to maximize these two log-likelihood functions using real data from \cite{birnbaum}. The data are fatigue time of 101 6061-T6 aluminum coupons cut parallel to the direction of rolling and oscillated at 18 cycles per second.

Informed by the results found from SAS, the first three parameters $(\alpha,\beta,s)$ are initialized over $U[0,200]$ and the last two, $(k,c)$, are initialized over $U[0,30]$ because they are shape parameters that are not supposed to be large. Positivity constraints apply to all elements of a particle, as required by definition. Particle elements are re-cast near zero ($U[0,0.5]$) if they stray negative. There were isolated instances of particles moving into positions yielding numerical errors in the likelihood computation (such as -$\infty$, $\infty$ or imaginary numbers). As a temporary way around this issue, particles are contained with an upper limit of 200. Elements are re-cast over $U[100,200]$ if the move beyond 200. It turns out in general, this is not a crucial requirement if numerical errors in the objective function evaluation are assigned the largest possible negative value.

Our first set of simulation results using PSO with 1000 particles at 2000 iterations for WBXII is shown in the \emph{fourth} row of Table \ref{table:wbxii}. The first PSO run for BBXII is also in the fourth row of Table \ref{table:bbxii}. These are noticeably different estimated coefficients compared to SAS (nlmixed) or R (nlminb and optim). The subsequent rows of Table \ref{table:wbxii} use a different way of initializing PSO particles.

Recast PSO \cite{jain2018review} is an enhanced version of standard PSO that introduces additional mechanisms to improve exploration and exploitation. It incorporates a memory mechanism to store the best positions found so far, called the Archive. The Archive maintains a collection of diverse and high-quality solutions discovered during the optimization process. Recasting operators are introduced to guide particles toward the solutions in the Archive. Recasting operators provide a way to dynamically adjust particle velocities based on the information stored in the Archive. By utilizing the Archive, Recast PSO aims to strike a balance between exploration and exploitation, potentially improving the convergence and diversity of solutions. \textit{Recast} PSO initializes each particle element within a neighborhood of the previous result so that the position can be further exploited. For example, to recast the $4^{th}$ row in the Table \ref{table:wbxii}, the first element of particles would be initialized over $(1-U[-0.1,0.1])\cdot138.96$, and the second element $(1-U[-0.1,0.1])\cdot0.9214$ and so on. For recasted PSO's, the originally imposed upper bound constraint is removed because particles are initialized to continue their previous exploitation phase and there is no exploration phase that may give some erratic particles that cause numerical errors in the likelihood computation.

The main motivation for recasting PSO runs is to squeeze out a higher precision in the maximized log-likelihood compared to the initial PSO by allowing an extended exploitation phase. The same exercise is done for BBXII in Table \ref{table:bbxii}, except that the first PSO run initialized particles over a more larger space of $U[0,400]$ and the last recast PSO hastily ``skipped ahead'' by initializing the first, third and fifth particle elements ($\alpha, s, c$) in a small neighborhood of (0.9268, 141.583, 0.027), and $\beta$ in $U[50,100]$ and $k$ in $U[0,0.05]$.

\begin{table*}[t]
	\centering
	\scalebox{0.82}{
		\begin{tabular}{|c c >{\centering\arraybackslash}p{2cm} c c c c c >{\centering\arraybackslash}p{2cm}|}
			\hline
			Method & Swarm Size & Number of Iterations & $\alpha$ & $\beta$ & $s$ & $k$ & $c$ & Log-Likelihood\\
			\hline
			NLMINB   &        &       &   0.278    &  3.5374 &  59.002  &  0.0018         & 554.947    & -458.298    \\
			OPTIM   &        &      & 18.077  & 0.8038733 &144.74  & 0.0513 & 11.156 & -455.0961    \\
			NLMIXED   &        &       &   70.84     &  0.8482  &  145.18  &  0.0126         & 10.638    & -455.09199    \\
			PSO           &  1000  & 2000  &   138.96     &  0.9214  &  143.73  &  0.0082         &  9.9763   & -455.09719    \\
			PSO (recast)  &  1000  & 2000  &   672.5      &  0.8597  &  145.26  &  $9.93\ast10^{-4}$  &  10.513   & -455.09113    \\
			PSO (recast)  &  1000  & 2000  &  38417.2     &  0.8653  &  145.26  &  $9.68\ast10^{-6}$  &  10.455   & -455.09099    \\
			PSO (recast)  &  1000  & 2000  &   106321.7   &  0.8653  &  145.26  &  $2.98\ast10^{-6}$  &  10.455   & -455.09099    \\
			\hline
		\end{tabular}
	}
	\caption{Sequential PSO estimates of WBXII.}
	\label{table:wbxii}
\end{table*}

\begin{table*}[t]
	\centering
	\scalebox{0.82}{
		\begin{tabular}{|c c >{\centering\arraybackslash}p{2.5cm} c c c c c >{\centering\arraybackslash}p{2.5cm}|}
			\hline
			Method & Swarm Size & Number of Iterations & $\alpha$ & $\beta$ & $s$ & $k$ & $c$ & Log-Likelihood\\
			\hline
			
			NLMINB   &        &      & 13.462 & 800.47 & 6950.562 &  10.8989   & 1.644 & -456.1897   \\
			OPTIM   &        &     & 0.9925 & 69.1949 & 142.1098 &  0.0229 &  9.3905 &  -455.1083    \\
			NLMIXED       &        &       &   300.03  &  96.98 &  140.2 & 2.083  & 0.51 & -457.193       \\
			PSO               &  1000  & 2000  &   1.8817	& 24.044	& 148.698	& 0.1537	& 6.010	& -455.34831   \\
			PSO (recast)      &  1000  & 2000  &   0.9268	& 23.350	& 141.584	& 0.06134	& 9.887	& -455.104862   \\
			PSO (recast)      &  1000  & 2000  &   0.9268	& 36.581	& 141.582	& 0.03913	& 9.888	& -455.104859  \\
			PSO (recast)  	   &  1000  & 2000  &   0.9267	& 40.416	& 141.582	& 0.03541	& 9.888	& -455.104858  \\
			PSO (recast)      &  1000  & 2000  &   0.9268	& 52.143	& 141.582	& 0.02744	& 9.888	& -455.104858  \\
			PSO (recast)$\ast$  &  1000  & 2000  &   0.9268	& 88.661	& 141.583	& 0.01613	& 9.887	& -455.104857  \\
			\hline
		\end{tabular}
	}
	\caption{Sequential PSO estimates of BBXII.}
	\label{table:bbxii}
\end{table*}

Between Table \ref{table:wbxii} and Table \ref{table:bbxii}, observe that as the maximized log-likelihood increases in decimal precision, some of the parameter estimates stabilize while some seem to diverge. For WBXII, each result of the recast PSO gave an increased estimate for $\alpha$ and decreased estimate for $k$. For BBXII, the divergent parameters are  $\beta$ and $k$. In either case, PSO finds more optimal solutions compared to SAS or R.

\begin{figure}
\centering
\includegraphics[width=8cm]{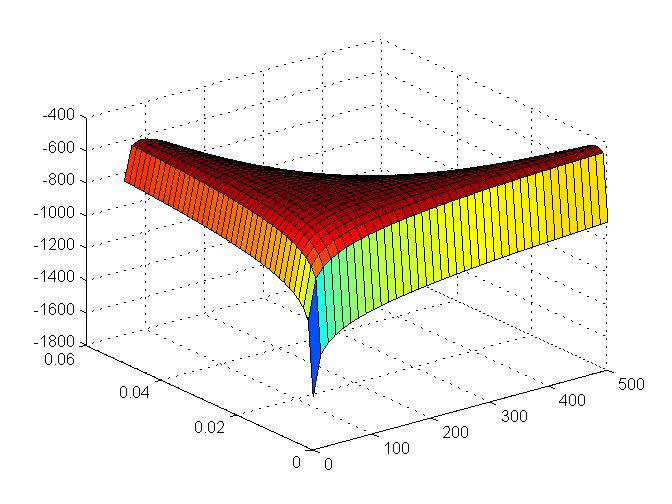}
\captionof{figure}{Surface of BBXII Log-Likelihood Profile with Fixed $(\alpha,s,c)$.}
\label{fig:bbxii_surface}	
\end{figure}

To make sense of why this happens and whether these estimates diverge without a bound, we inspect the profile of the likelihood function where the ``stable'' parameters are fixed at their nominal estimates in Table \ref{table:wbxii} and Table \ref{table:bbxii}. Figure \ref{fig:bbxii_surface} shows the apparent flat spot in the likelihood profile as  $\beta$ and $k$ diverge from each other. We found that in fact, $\beta$ can be arbitrarily large for a small enough $k$. This clearly implies that the parameter pair are unidentified given the data. The picture for WBXII is identical.

The SAS outputs of the results for WBXII and BBXII were generated normally, without a hint of any problem with the estimation procedure. An inspection of the log file did reveal a general warning. Note that the final Hessian matrix is full rank but has at least one negative eigenvalue. Second-order optimality condition violated. We specified a variety of different convergence criterion in the SAS routine and were unable to improve on the maximized log-likelihood shown in the tables and unable to pick up any divergent patterns of parameter estimates as seen in the tables. Using SAS, it is difficult to gain any insight to which parameters of the model are unidentified or redundant.

The use of R routines proved to be a similar experience to SAS. The results presented in the tables are the best that can be achieved by adjusting the routine options. For BBXII using nlminb, the routine gives no warning message and falsely concludes a convergent result. For other R exercises, the routines give a warning message that is unrelated to Hessians or parameter identification, but rather that the iterations ran into numerical errors in the evalution of the log likelihood. Perhaps better results could be obtained with R if these numerical errors can be bypassed, but there is little the user can do because these are canned routines.

With PSO, we were able to isolate the unidentified parameters with a series of exploitative recast PSO runs. It became quickly apparent after trying different swarm size and iteration combinations that there were stable as well as diverging coefficients. In fact, the series of recast PSO's are only reported here to illustrate the investigative process -- the divergent nature of the estimates on the unidentified pairs of parameters are able to be seen in a single PSO run with a large swarm size and many iterations, where particles are initialized in the same way as above, but without upper bound re-randomization. However, this method was slow because a large swarm size slows down the implementation in MATLAB. This would not pose a practical problem, however, if PSO is implemented in C++ or Rcpp.

\section{Log Binomial Regression}
\subsection{Convergence Failures in Log Binomial Regressions}
\label{sec4}
\begin{figure}
\centering
\includegraphics[width=13cm]{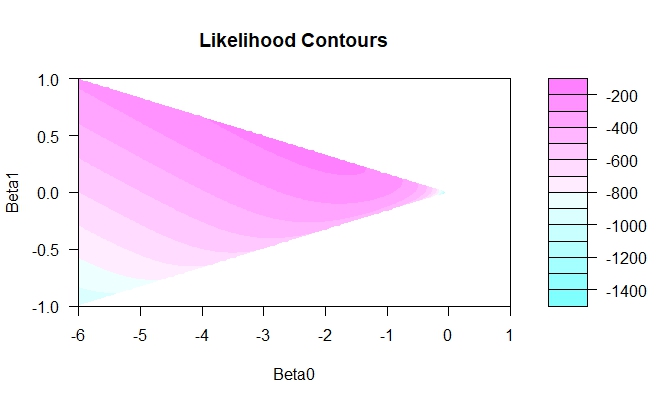}
\captionof{figure}{Surface of Log Binomial Regression Likelihood with One Intercept and Covariate.}
\label{fig:logbinom_surface}
\end{figure}

As noted in \cite{hosmer}, estimates of the relative risk ratio, controlling for confounders, can be obtained from a fitted logistic regression. Unfortunately, this will be over estimated when the outcome is not rare. The log binomial regression is desirable to researchers because the relative rick can be directly estimated from its coefficients even when the outcome is not rare. However, it is notoriously common in practice that the log binomial regression results in convergence failures using the iterative methods commonly offered in statistical software \cite{williamson}. This is because the log link function, unlike say, the logit link, can yield greater than 1 probabilities if the parameter space is not constrained. Problems can further arise if the optimum is near the boundary of the constrained space. Figure \ref{fig:logbinom_surface}, shows the contours of the likelihood using one instance of our simulation data.

For the data in this section, we follow the design in Blizzard and Hosmer (2006), in particular setting 1 of Table 1. The true parameter values are $\beta_0 =-2.30259$ and $\beta_1 =0.38376$. The covariate is uniformly distributed over [-6,6]. The authors, who used STATA, report two types of adverse estimation outcomes. The iterative method used by the software may fail to converge, or a converged solution may not be admissible. An admissible solution $\boldsymbol{\hat{\beta}}$ is defined to satisfy $\pi(\mathbf{x}_i)=\exp(\mathbf{x}_i '\boldsymbol{\hat{\beta}}) <1$ for all $i$, for the vector of covariates $\mathbf{x}_i$ of the $i^{th}$ observation.  In R glm() however, all solutions are admissible because the quasi-Newton steps are truncated if the iteration goes out of bounds. Using the more general purpose optim(), the hand written log likelihood function can be set to equal a large penalty if the parameter is inadmissible. This is commonly done whenever the parameter space needs to be constrained. This method however, provided consistently poor results compared to the built in glm() function in R.

After simulating a 1000 data sets, using initial values of $(\beta_0, \beta_1)=(-0.1, 0)$ (because glm() would not even begin to run with $(0,0)$), we found that over $41\%$ of times the warning was issued for "non-convergence" using R glm(). If the initial value used was closer to the true value, then the non-convergence percentage would decrease to about $30\%$. Taking one of these non-convergent samples, we ran the iterative procedure over a full array of initial values to see if any convergence can be achieved at all. Figure \ref{fig:startingpoints} plots the convergent initial values when using a data set which was non-convergent using the initial value of (-0.1, 0). For a convergent sample, all initial values over the admissible region resulted in convergence - Figure \ref{fig:startingpoints} would look like a black isosceles triangle.

\begingroup
\centering
\includegraphics[width=13cm]{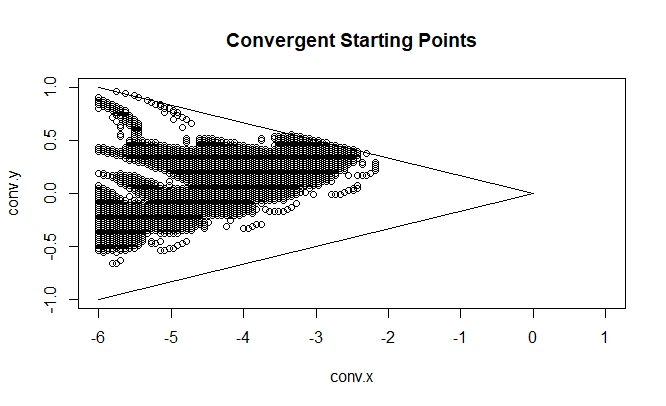}
\captionof{figure}{Convergent Initial Values for some Non-Convergent Sample.}
\label{fig:startingpoints}	
\endgroup

When comparing PSO against any convergent result from R glm(), the results were identical. Our interest naturally turned to how PSO estimates compare when R glm() estimates were deemed non-convergent. It turns out that even when the outcome is not convergent, the estimated coefficients and maximized log-likelihood of R glm() are not terrible. Using a starting value of (-0.1, 0), we generated samples until we collected 200 non-convergent samples. For each of these samples we compare the results of R glm() and PSO. The loglikelihood to be maximized by PSO is

\begin{align}
\sum^n_{i=1} \left[y_i (\beta_0+\beta_1 x_{1i})+(1-y_i)\ln(1-\exp{\beta_0+\beta_1 x_{1i}})\right]  .
\end{align}

The swarm size was 200 and 1500 iterations were used. Particles were uniformly initialized over a [-3,3] box. When comparing the 200 results of both methods, 100\% of time, PSO resulted in a higher log likelihood. On average better by 0.37, with a standard deviation of 2.3. Sometimes the difference was as much as 32. The average maximized log likelihood is around -164. On average, the estimate for $\beta_0$ differed by 0.07 with a standard deviation of 0.08, sometimes differing by as much as 0.6. For $\beta_1$, they differed on average by 0.01 with a standard deviation of 0.02, and a maximum difference of 0.22. The average relative bias (computed as $100*\frac{1}{200} \sum^{200}_{i=1} (\hat{\beta_{0i}}-\beta_{0i})/\beta_{0i}$) of $\beta_0$ and $\beta_1$ for PSO was 1.37 and 1.73 respectively, while for R glm() it was -2.26, -2.16. If we compared PSO to optim(), which is the go-to tool in applied statistics work for custom made objective functions, these comparisons would be more dramatic in favor of PSO.

In practice, on real world data sets with higher dimensions, finding a suitable set of initial values that fall in the admissible parameter space can be hard for the practitioner without trying brute force methods.  Even within the admissible parameter space, the data maybe such that the set of convergent initial values is sparse in an extreme version of Figure \ref{fig:startingpoints}. Often, practitioners have to settle for alternative models such as Poisson regression or data augmentation schemes to obtain an estimate of the relative risk \cite{hosmer} \cite{williamson}. PSO bypasses the issue of constrained parameter space, initial value selection and non-convergence. When the outcome of an iterative method is non-convergent we also showed that PSO always finds more optimal estimates.

\subsection{Prediction of Heart Disease Using Penalized Log Binomial Regression}

According to the World Health Organization, approximately 12 million people worldwide die from heart diseases each year. Cardiovascular diseases account for half of these deaths in developed countries, including the United States. Identifying the risk factors of heart diseases early on can help patients make informed decisions about lifestyle changes and ultimately reduce the risk of complications. This research aims to use regression models to identify the most significant risk factors of heart disease and predict the overall risk. The data set used in this study is publicly available on Kaggle (\url{https://www.kaggle.com/datasets/dileep070/heart-disease-prediction-using-logistic-regression}, accessed on 05/26/2023) and is based on an ongoing cardiovascular study conducted on residents of the town of Framingham.The objective of the classification is to anticipate if a patient has the possibility of developing coronary heart disease (CHD) within the next ten years. The data set contains information about the patients and comprises of 15 attributes, with more than 4,000 records. In our data analysis, we discard the observations with missing values, resulting in a data matrix with 3656 observations (1622 males and 2034 females).

Suppose we are interested in estimating relative risk of developing CHD between male and female groups. Table~\ref{tab:UKB} summarizes the covariates and the response (Ten Year CHD) used in our log binomial regression model. Instead of the logistic regression, log binomial regression is more straightforward to apply in this case \cite{hosmer}. Further, to reduce the variance of the estimator, we add a $L_1$ regularization term to the likelihood function so that we have a LASSO optimization problem \cite{tibshirani1996regression}. To sum up, suppose we observe $n$ iid variables $(Y_i, X_i),i=1,2,\cdots,n$ and $Y_i|X_i=x_i\sim \mathcal{B}er(p_i)$ where $\log p_i=x_i^T\beta$. The log-likelihood function for $\beta$ is
$$l(\beta)= \sum_{i=1}^n\left(y_ix_i^T\beta + (1-y_i)\log(1-e^{x_i^T\beta})\right)$$
with the obvious constraint
$$x_i^T\beta\le 0$$
for all $i$. To extend it to binomial regression, assume that $Y_i|X_i=x_i\sim \mathcal{B}in(n_i, p_i)$ where $\log p_i=x_i^T\beta$. The log-likelihood function for $\beta$ is
$$l(\beta)= \sum_{i=1}^n\left(\log{n_i \choose y_i} + y_ix_i^T\beta + (n_i-y_i)\log(1-e^{x_i^T\beta})\right)$$
with the same constraint. The corresponding LASSO optimization problem is
\begin{align}\label{eq:log_bin_lik}
    &\min_{\beta}\ -l(\beta) +\rho\lVert \beta\lVert_1\\
    &\text{such that }x_i^T\beta\le 0,\ i=1,\cdots,n.\nonumber
\end{align}
where $\rho$ is the regularization parameter. However, if the sample size $n$ becomes large, it is difficult for a numerical algorithm to satisfy the conditions $x_i^T\beta\le 0, i=1,\cdots,n$ simultaneously. Hence, we rewrite the constraints and $\log p$ into an alternative form:
$$\log p=\min(x_i^T\beta,\ 0).$$
Such modification can avoid numerical issues when we are applying PSO to solve for $\beta$. But we also note that it leads to a different optimization problem compared with formulation (\ref{eq:log_bin_lik}): the parameter space of the former is $\mathbb{R}^d$ where $d=\text{dim}(X_i)$ while the latter is a sub-manifold of $\mathbb{R}^d$. We implemented the PSO codes for penalized log-binomial regression in Python via the package \emph{pyswarms} \cite{miranda2018pyswarms}. 

\begin{table}[!htbp]
\caption{Baseline characteristics of a subgroup within the World Health Organization (WHO)}
\centering
\renewcommand{\arraystretch}{.8}
\begin{tabular}{lrrr}
\hline \hline
    Characteristic & Female (n=2,419) & Male (n=1,819) & p-value \\
\hline
    Age, mean $\pm$ SD, year & 49 $\pm$ 6 & 48 $\pm$ 6 & 0.054 \\
    Education\\
    \,\,\,\,\,\,\,\,\,\,1 & 943 (40\%) & 777 (44\%) \\
    \,\,\,\,\,\,\,\,\,\,2 & 764 (32\%) & 489 (28\%) \\
    \,\,\,\,\,\,\,\,\,\,3 & 463 (20\%) & 224 (13\%) \\
    \,\,\,\,\,\,\,\,\,\,4 & 197 (8\%) & 276 (15\%)\\
    \,\,\,\,\,\,\,\,\,\,Unknown & 52  & 53 \\
    
Smoking status &       &  & $<$0.001\\
    \,\,\,\,\,\,\,\,\,\,Current smoker & 988 (41\%)  &  1,106 (61\%)\\
    \,\,\,\,\,\,\,\,\,\,Nonsmoker & 1431 (59\%)   & 713 (39\%) \\
    BMI, mean $\pm$ SD &   24.8 $\pm$ 2.3    & 26.1 $\pm$ 2.1 & $\le0.001$ \\
   \,\,\,\,\,\,\,\,\,\,Unknown   & 14  & 5 \\
   Heart rate, mean $\pm$ SD & 75 $\pm$ 6 & 75 $\pm$ 9 & $\le0.01$
   \\
   Cigarettes Per Day, 95\% quantile & 0 (0, 10) &   \\
   Prevalent Stroke  & 15 (0.6\%) & 10 (0.5\%) & 0.8\\
   Prevalent Hyp & 746 (31\%) & 570 (31\%) & 0.7 \\
   Total cholesterol level, mean $\pm$ SD & 237 $\pm$ 31 & 230 $\pm$ 24 & $<0.001$ \\
   \,\,\,\,\,\,\,\,\,\,Unknown   & 43  & 7 \\
    Diabetes & & & 0.3\\
    \,\,\,\,\,\,\,\,\,\,Yes & 57 (2.4\%) & 52 (2.9\%) \\
    \,\,\,\,\,\,\,\,\,\,No & 2362 (97.6\%) & 1767 (97.1\%)\\
    Glucose level, mean $\pm$ SD & 78 $\pm$ 6 & 78 $\pm$ 8 & 0.6  \\
    \,\,\,\,\,\,\,\,\,\,Unknown   & 273  & 115 \\
    Ten Year CHD & & & 	$<$0.001 \\
    \,\,\,\,\,\,\,\,\,\,Yes & 301 (12\%) & 343 (19\%) & \\
    \,\,\,\,\,\,\,\,\,\,No  & 2118 (88\%) & 1476 (81\%) & \\
\hline \hline
\end{tabular}
\vspace{.3cm}
\begin{tablenotes}
\item 
    \item BMI $=$ body mass index; CHD = coronary heart disease; p-values are calculated using Wilcoxon rank sum test and Pearson's Chi-squared test.
    \end{tablenotes}
\label{tab:UKB}
\end{table}

\begin{figure}
    \centering
    \includegraphics[scale=0.6]{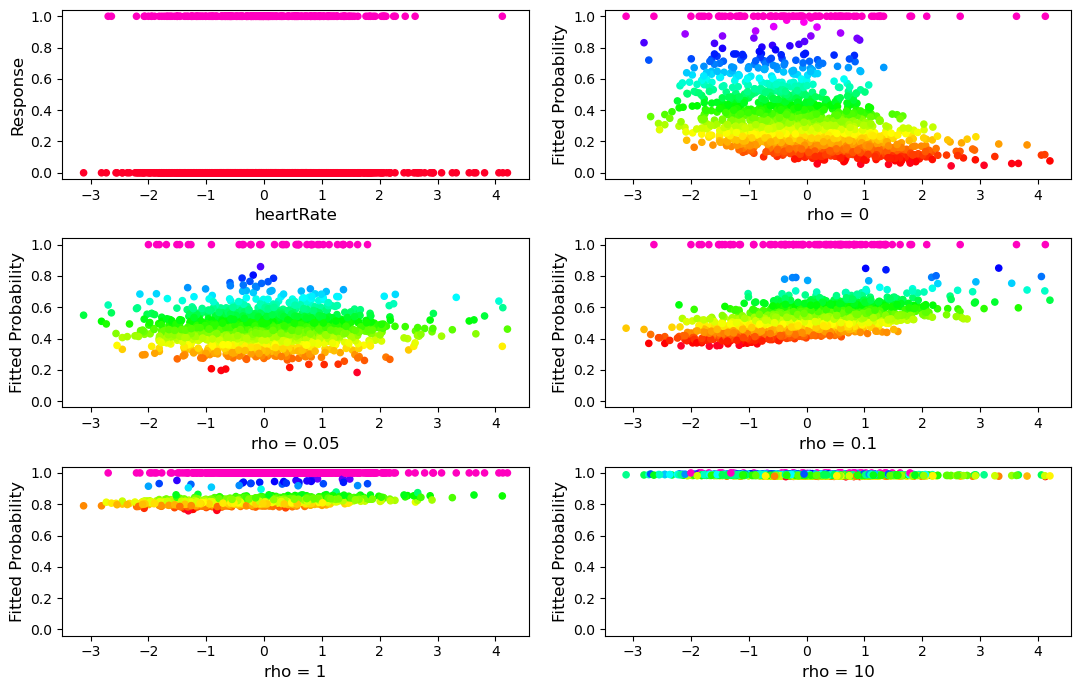}
    \caption{\textcolor{black}{Fitted probabilities of log binomial LASSO regression for the male cohort. Various shades indicate distinct fitted probability values - red indicating a relatively low probability. Each data point corresponds to a unique observation, totaling 1622 male participants. As we increase $\rho$, the fitted probabilities become more clustered around higher values which aligns with our objective function penalty term. Since a higher $\rho$ will penalizes $\beta$ towards zero,the predicted probabilities tend to one.}}
    \label{fig:log_bin_male}
\end{figure}

\begin{figure}
    \centering
    \includegraphics[scale=0.6]{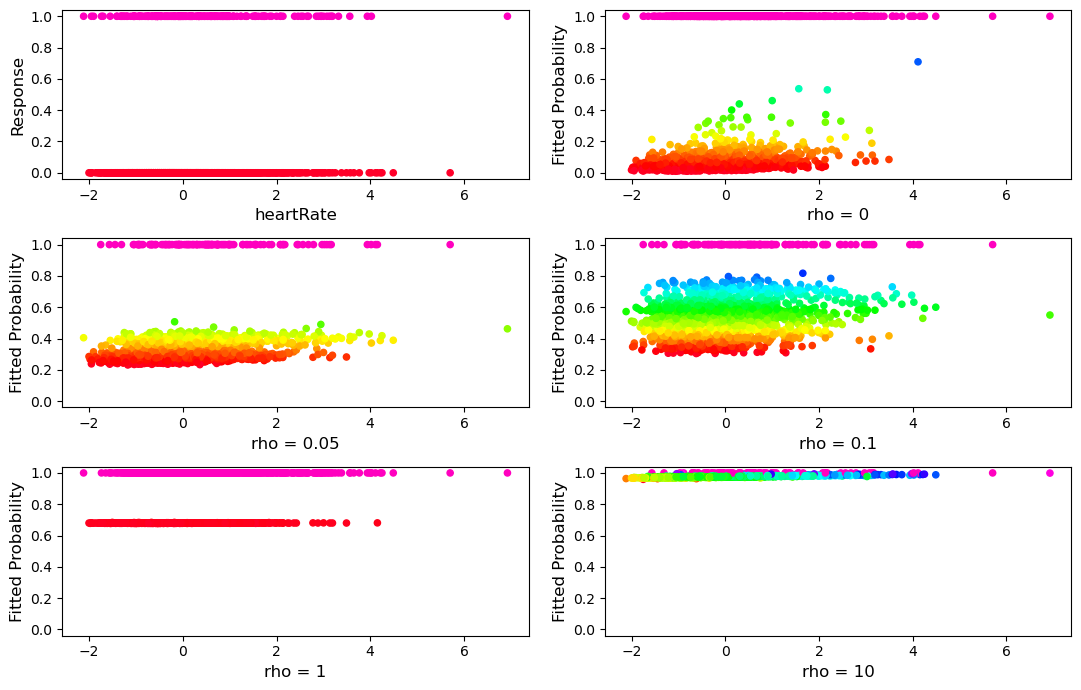}
    \caption{\textcolor{black}{Fitted probabilities of log binomial LASSO regression for the female cohort. Similar to that of the male cohort, the female cohort's predicted probabilities cluster around higher values as we increase $\rho$. The dots represent the 2034 non-missing females.}}
    \label{fig:log_bin_female}
\end{figure}

We use the global best PSO to estimate parameters in the LASSO log binomial regression problem with $\rho\in\{0, 0.05, 0.1, 1, 10, 100\}$ (note that $\rho=0$ corresponds to the maximum likelihood estimation in a log binomial regression). The hyper-parameters of PSO are set to $c_1=0.5,\ c_2=0.3,\ w=0.9$ and we run the algorithm with 300 iterations for each $\rho$. The original data and fitted / estimated probability of the WHO data set within the male and female cohorts are shown in figure~\ref{fig:log_bin_male} and \ref{fig:log_bin_female} with the x-axis refers to standardized heart rate (one of the covariates used in our model, see table~\ref{tab:UKB}). From the figure we see that as $\rho$ becomes larger, the estimated probability tends to $1$, and when $\rho=0$, the estimated probability is more similar to the original data (top left panel) resulting in overfitting. We illustrate the estimated parameters with $\rho=0.1$ in table~\ref{tab:log_bin} (the choice $\rho=0.1$ is selected via a 5-fold cross-validation). The model selects Age, BPMeds, Prevalent Hyp, Diabetes, Systolic Blood Pressure, Diastolic Blood Pressure and Glucose for the male cohort and Age, BPMeds, Systolic Blood Pressure for the female cohort. There are three different algorithms in the table. The first two are global and local best PSO algorithms \cite{kennedy} while the third one is the competitive swarm optimizer with mutated agents \cite{cheng2014competitive, zhang2020competitive}. It is obvious that both PSOs outperforms CSOMA in terms of best likelihood values and estimates consistency.

\begin{table}[]
    \centering
    \begin{tabular}{lccc}
    \hline\hline
    \multicolumn{4}{c}{The Male Cohort (n=1622)}\\
    \hline
       \textbf{Algorithm} & \textbf{Global PSO} & \textbf{Local PSO} & \textbf{CSOMA} \\
       \hline         \textbf{Best value} & 2376.231 & 2378.930 & 2376.634\\
        \hline
       Age  & 0.043 & 0.036 & 0.044\\
        Education & 0.000 & 0.000&0.000 \\
       Smoking status & 0.000 & 0.000&0.011\\
        Cigarettes Per Day  & 0.000 &0.000 &0.000\\
        BPMeds & 0.000 &0.000&0.000\\
         Prevalent Stroke & 0.000 &0.000&0.000\\
        Prevalent Hyp & 0.000 & 0.000&0.000 \\
        Diabetes & 4.109 & 4.092 &3.951\\
        Total Cholesterol & 0.000 & 0.000 &0.000\\
        Systolic Blood Pressure & 0.047 & 0.044&0.036 \\
        Diastolic Blood Pressure & 0.000 & 0.000&0.010\\
        BMI & 0.000 &0.000&0.000\\
        Heart Rate & 0.000 & 0.000&0.000\\
        Glucose & 0.000 & 0.000 &0.010\\
       \hline\hline\\
       \\
    \hline\hline
    \multicolumn{4}{c}{The Female Cohort (n=2034)}\\
    \hline

       \textbf{Algorithm} & \textbf{Global PSO} & \textbf{Local PSO} & \textbf{CSOMA} \\
       \hline
        \textbf{Best value} & 3070.909 & 3084.967 & 3267.877\\
        \hline
       Age  & 0.030 & 0.025 & 0.019  \\
        Education & 0.000 & 0.003 & 0.000 \\
       Smoking status & 0.000 & 0.000 & 0.000 \\
        Cigarettes Per Day  & 0.000 & 0.000& 0.000 \\
        BPMeds &  0.000& 0.000 & 4.270 \\
         Prevalent Stroke & 0.000& 0.000  &0.000\\
        Prevalent Hyp & 0.000 & 0.000 &0.012 \\
        Diabetes &4.793 & 4.889&-1.008\\
        Total Cholesterol & 0.000 & 0.012 &0.000 \\
        Systolic Blood Pressure & 0.020& 0.017&0.040 \\
        Diastolic Blood Pressure & 0.000 & 0.000 &0.000\\
        BMI & 0.000& 0.002 &0.000\\
        Heart Rate & 0.000 & 0.001&0.000 \\
        Glucose & 0.000 & 0.000&0.000\\
       \hline\hline\\
    
    \end{tabular}
    \caption{\textcolor{black}{Parameter estimation in the log binomial regression model using the WHO data set with $\rho=0.1$, tuned by 5-fold cross validation. The best value indicates the best negative log-likelihood value, obtained from 50 runs. Therefore, a smaller best value indicates a better result. Global PSO = Global Best PSO; Local PSO = Local Best PSO; CSOMA = Competitive Swarm Optimizers with Mutated Agents.}}
    \label{tab:log_bin}
\end{table}

To further investigate the speed of PSO and CSOMA applied to log binomial regression, we compare it with the standard package \textcolor{black}{\emph{logbin}} in R.  It is worth noting that Stata glm command only supports binomial regression with a logit link instead of a log link. SAS PROC Genmod doesn't support the log link for binomial regression either. The package \emph{logbin} implemented by \cite{donoghoe2018logbin} uses the combinatorial expectation maximization (CEM), a variant of the EM algorithm proposed by \cite{marschner2012relative, marschner2014combinatorial}. Unfortunately, this package does not implement penalized log binomial regression but the usual maximum likelihood estimation, so we set $\rho=0$ in the optimization problem~\ref{eq:log_bin_lik}. Further, they also implemented the conventional EM algorithm \cite{dempster1977maximum} in the \emph{logbin} package and we compare it with PSO and CSOMA. We run each algorithm in both the female and the male cohort 50 times to get reasonable statistical results and the results (estimates, time and standard deviation) are summarized in table~\ref{tab:comp_PSO_CEM}. Note that the best value refers to the log likelihood function in equation~\ref{eq:log_bin_lik}. The number of iterations of PSO and CSOMA are both set to 800, the number of neighbors of local best PSO is set to 6, and the EM are set to 15 iterations. For the CEM algorithm, the running time is too long to be recorded (more than 3 minutes) and it will be meaningless to record the running time so we only keep track of its best values (see table~\ref{tab:comp_PSO_CEM}).  

\begin{table}[]
    \centering
    \begin{tabular}{lccccc}
    \hline\hline
     \multicolumn{6}{c}{The Male Cohort}    \\
    \hline
    \textbf{Algorithm} & \textbf{Global PSO} & \textbf{Local PSO} & \textbf{CSOMA} & \textbf{CEM} & \textbf{EM} \\
    \hline
    \textbf{Time (SD)} & 2.692 (0.082)  &  3.275 (0.077) &3.132 (0.051) &too long & 0.0196 (0.006)  \\
    \hline 
     \textbf{Mean value (SD)} & 1012.861 (0.132) & 1015.940 (0.546) &1313.991 (5.481) & 1397.077 (0.000)& 1235.448 (0.001)  \\
         \hline\hline\\
         \hline\hline
     \multicolumn{6}{c}{The Female Cohort}    \\
    \hline
    \textbf{Algorithm} & \textbf{Global PSO} & \textbf{Local PSO} & \textbf{CSOMA} & \textbf{CEM} & \textbf{EM} \\
    \hline
    \textbf{Mean Time (SD)} &  3.012 (0.062) & 3.619 (0.055) & 3.476 (0.045) & too long &0.0292 (0.009)  \\
    \hline 
     \textbf{Mean value (SD)} & 1096.242 (0.150) & 1119.541 (0.164) & 1492.033 (90.23) &1195.444 (0.000) & 1328.252 (0.001)  \\
     \hline\hline
    \end{tabular}
    \caption{\textcolor{black}{Comparison of PSO and CEM: when $\rho=0$, tuned by 5-fold cross validation. The mean value represents the average negative log likelihood values obtained from the 50 runs. Thus, a smaller value indicates a better performing algorithm. The mean time indicates the average time it takes to complete running one round of the specific algorithm listed. We can see Global PSO outperforms both the Local PSO and CSOMA in terms of the two aforementioned metrics for both cohorts. For both cohorts, even though Global PSO under-performs the EM algorithm in terms of mean running time, it outperforms EM in terms of mean negative log likelihood by a significant amount. Furthermore, the CEM algorithm, despite achieving the smallest mean log likelihood among all five algorithms, the mean time it takes to complete one run is more than three minutes, making it impractical to scale to larger data sets or data sets with a large number of parameters.}}
    \label{tab:comp_PSO_CEM}
\end{table}

\section{Maximum Likelihood Estimation in the Exponentiated Exponential-Inverse Weibull (EE-IW) Model}\label{sec:mle_eeiw}

\textcolor{black}{The Exponentiated Exponential-Inverse Weibull (EE-IW) model is a statistical distribution commonly used in survival analysis and reliability modeling to represent survival or lifetime data. It is a flexible parametric model that combines the exponential distribution and the inverse Weibull distribution. The model allows for different shapes of hazard functions, which describe the instantaneous risk of an event occurring. The EE-IW model is particularly useful because it can capture a wide range of survival patterns. It can accommodate decreasing hazards (where the risk of an event decreases over time), constant hazards (where the risk remains constant), and increasing hazards (where the risk increases over time). This flexibility makes it applicable to various scenarios in survival analysis, such as studying the lifetimes of mechanical components, analyzing medical survival data, or examining the reliability of systems.To estimate the parameters of the EE-IW model, statistical techniques such as maximum likelihood estimation (MLE) are commonly employed. MLE involves finding the parameter values that maximize the likelihood of the observed data given the model. Once the parameters are estimated, the model can be used to make predictions about future lifetimes or analyze the survival characteristics of a population. The probability density function (PDF) of the EE-IW distribution, as given in equation (9), consists of a normalization constant, shape parameters (c, $\alpha, and\  \beta$), and two component functions (g(x) and G(x)). The cumulative distribution function (CDF) of the EE-IW distribution is shown in  (12) and it represents the probability that a random variable that follows the EE-IW distribution  is less than or equal to a given value x. In order to evaluate the goodness of fit and compare different estimation methods, the log-likelihood function (equation 13) is used. The log-likelihood function quantifies how well the estimated model fits the observed data, with higher likelihood values indicating better fit. 
The pdf is \cite{badr2022exponentiated}
\begin{align}
    f(x)&=c\alpha\beta g(x)[G(x)]^{c-1}(1-G^{c}(x))^{\beta-1}
\end{align}
where $c,\alpha,\text{ and }\beta$ are the shape parameters, and 
\begin{align}
    g(x)&=\alpha x^{-\alpha-1}e^{-x^{-\alpha}};\ \ x>0,a>0\\
G(x)&=e^{-x^{-\alpha}};\ \ x>0,a>0.
\end{align}
The corresponding cdf for the EE-IW is 
\begin{align}
    F(x)&=1-[1-(e^{-cx^{-a}})]^{\beta};\ \ \beta,\alpha,c,x>0
\end{align}
It follows that the log-likelihood function is
\begin{align}  l(\alpha,\beta,c)&=n\ln{c}+n\ln{\alpha}+n\ln{\beta}-(\alpha+1)\sum_{i=1}^{n}\ln{x_{i}}-c\sum_{i=1}^{n}{x_i}^{-\alpha}+(\beta-1)\sum_{i=1}^{n}\ln{[1-(e^{-c{x_{i}}^{-\alpha}})]}
\end{align}
 the MLEs for the parameters are given by 
$$\widehat{\alpha},\widehat{\beta},\widehat{c}=\arg\max_{\alpha,\beta,c}=\text{l}(\alpha,\beta,c).$$
The specific details and mathematical steps involved in deriving the maximum likelihood estimate $\widehat{\beta}_{MLE}$ for the Exponentiated Exponential-Inverse Weibull (EE-IW) model is explained in \cite{almetwally2021extended}. 
We compare our PSO MLE estimates with the ones reported in \cite{badr2022exponentiated} using the EE-IW distribution proposed in \cite{elbatal2014exponentiated}.To evaluate and compare the performance of different estimation methods, the Cumulative Distribution Function (CDF) is estimated using PSO, Badr and Sobahi's method, and the empirical CDF (ECDF). We apply PSO using \textit{pyswarm} package \cite{miranda2018pyswarms} to find EE-IW MLEs using three real-world datasets, including a 32-day COVID-19 DATASETS by \cite{almetwally2021extended}, breaking stress of carbon fibers by \cite{nichols2006bootstrap}, and deep groove ball bearings endurance test by \cite{lawless2011statistical}.
\begin{table}[]
    \centering
    \begin{tabular}{c|l rrrr}\hline\hline
        Dataset & Method & Log Likelihood & $\widehat{\alpha}$ & $\widehat{\beta}$ & $\widehat{c}$ \\\hline
        1 & PSO & 95.371 & 1.232 (0.019) & 293.441 (0.445) & 0.271 (0.016) \\ 
         & Reported & 83.970 & 1.573 (0.043) & 5.431 (1.756) & 0.040 (0.005) \\
        2 & PSO & -141.889 & 0.410 (0.034) & 657.059 (0.129) & 9.969 (0.315)\\
         & Reported & -237.205 & 6.146 (0.167) & 0.108 (0.042) & 0.021 (0.009)\\
        3 & PSO & -113.461 & 0.694 (0.291) &16.537 (1.841) &57.825 (26.638) \\
         & Reported & -150.318 & 3.951 (0.501)&0.0641 (218.096)& 2.303 (29.337)\\
        \hline\hline
    \end{tabular}
    \caption{A comparison of MLE Estimates by PSO and Badr and Sobahi's method. PSO returns different but better results in terms of likliehood values and standard errors. The numbers inside the braces are standard error estimations based on Fisher information computed in Badr and Sobahi (2022).}
    \label{tab:EE-IW}
\end{table}
We further compare the cumulative distribution function (CDF) estimated by PSO, by Badr and Sobahi's method, and by empirical CDF, given by the formula: $\hat{F}(x) = \frac{1}{n} \sum_{i=1}^{n} \mathbf{1}(X_i \leq x)$. A closer alignment between the estimated CDF and the empirical CDF indicates a better fit of the model to the data. Based on the presented results, the PSO-based method outperforms Badr and Sobahi's method in all three datasets, indicating its superior performance in fitting the data to the EE-IW model. See Figure \ref{fig:dataset1}.
}

\textcolor{black}{
\newpage
\begin{figure}
    \centering
    \includegraphics[scale=0.7]{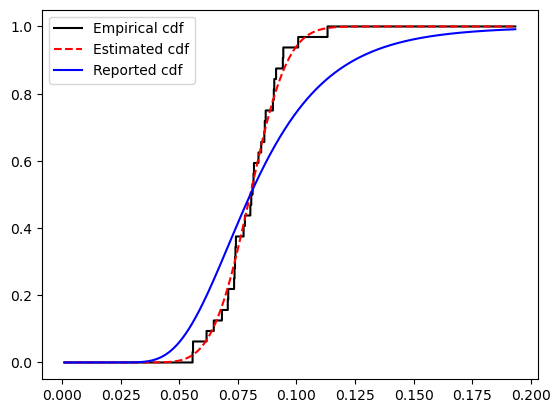}
    \includegraphics[scale=0.7]{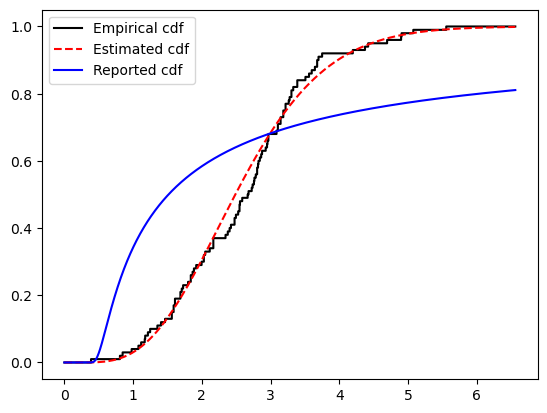}
    \includegraphics[scale=0.7]{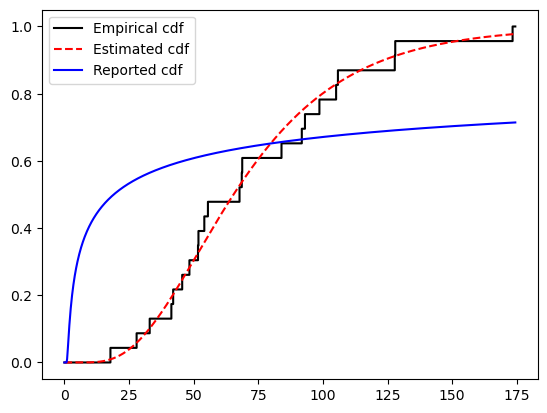}
    \caption{Fitted EE-IW model using three different datasets suggested in \cite{badr2022exponentiated}. Top: The COVID-19 data in \cite{almetwally2021extended}; middle: The breaking stress of carbon fibers (in Gba) data in \cite{nichols2006bootstrap}; bottom: The deep groove ball bearings data in \cite{lawless2011statistical}.The black line represents the empirical CDF. The blue line represents the estimated CDF by Badr and Sobahi's method. The red line represents the estimated CDF by PSO. We can see that the closer the estimated CDF is to the empical CDF curve, the better the methodology is. In this case, PSO outperforms Badr and Sobahi's method in all three datasets. }
    \label{fig:dataset1}
\end{figure}}

\section{Conclusion}
\label{sec5}
\textcolor{black}{Our research furnishes substantial evidence in favor of the Particle Swarm Optimization (PSO) algorithm as a reliable and robust method for optimizing objective functions in parameter estimation tasks. While popular software packages like SAS or R offer tried-and-true routines that perform effectively in numerous situations, there are certain research processes that present significant numerical challenges, especially those involving the WBXII and BBXII distributions. Traditional black-box routines may falter in these scenarios, lacking the requisite flexibility and transparency to manage the unique estimation problems that emerge. It is here that PSO steps in as a promising alternative. PSO stands out in handling complex cases, particularly when dealing with non-trivial constraints within the parameter space, situations where the optimum lies close to a boundary, or instances where the likelihood function asymptotes or features a flat region. Drawing on the swarm intelligence of particles and their collaborative exploration of the parameter space, PSO showcases its adaptability and efficiency in surmounting these intricate optimization challenges. Moreover, PSO's capacity to accommodate user-defined link functions in binomial regression settings sets it apart from conventional software packages like R and SAS. This adaptability enables researchers to customize the estimation process to their unique requirements and paves the way for a more in-depth analysis of intricate models. In our study of the EE-IW distribution, PSO consistently outstripped traditional statistical estimation methods reliant on moment-based approaches. This superior performance stems from PSO's skill in traversing the parameter space and pinpointing optimal solutions, thereby yielding more precise and dependable parameter estimates.
The accessibility of PSO further adds to its allure as a practical tool for parameter estimation. Even those without specialized optimization knowledge can effortlessly apply the PSO algorithm, thanks to available libraries and frameworks. The transparency inherent in PSO's iterative process and convergence behavior grants researchers a clear view into the estimation procedure, enhancing the interpretability of the findings. In conclusion, the PSO algorithm's versatility, adaptability, and transparency establish it as a fitting and potent technique for addressing diverse estimation challenges across various research domains. Its excellence in intricate optimization scenarios, alignment with user-defined link functions, and straightforward implementation enshrine PSO as a valuable asset. It is poised to enrich parameter estimation methodologies, equipping researchers with a powerful instrument to distill meaningful insights from their data. }

\bigskip
\begin{center}
{\large\bf SUPPLEMENTARY MATERIAL}
\end{center}
\begin{description}

\item[Codes:] Codes in R, SAS and MATLAB are available upon request.

\item[Data for Section \ref{sec2}:] $\mathbf{x}$=(0.55, 0.74, 0.77, 0.81, 0.84, 0.93, 1.04, 1.11, 1.13, 1.24, 1.27, 1.28, 1.29, 1.30, 1.36, 1.39, 1.42, 1.48, 1.49, 1.50, 1.50, 1.52, 1.53, 1.54, 1.55, 1.55, 1.58, 1.59, 1.60, 1.61, 1.61, 1.61, 1.61, 1.62, 1.62, 1.63, 1.64, 1.66, 1.66, 1.66, 1.67, 1.68, 1.68, 1.69, 1.70, 1.73, 1.76, 1.77, 1.78, 1.81, 1.82, 1.84, 1.84, 1.89, 2.00, 2.01, 2.24)$'$.

\item[Data for Section \ref{sec3}:] $\mathbf{x}$=(70, 90, 96, 97, 99, 100, 103, 104, 104, 105, 107, 108, 108, 108, 109, 109, 112, 112, 113, 114, 114, 114, 116, 119, 120, 120, 120, 121, 121, 123, 124, 124, 124, 124, 124, 128, 128, 129, 129, 130, 130, 130, 131, 131, 131, 131, 131, 132, 132, 133, 134, 134, 134, 134, 134, 136, 136, 137, 138, 138, 138, 139, 139, 141, 141, 142, 142, 142, 142, 142, 142, 144, 144, 145, 146, 148, 149, 151, 151, 152, 155, 156, 157, 157, 157, 157, 158, 159, 162, 163, 163, 164, 166, 166, 168, 170, 174, 196, 212)$'$.

\item[First Dataset for Section \ref{sec:mle_eeiw}:] $\mathbf{x}$=(0.0557, 0.0559, 0.0617, 0.0649, 0.0683, 0.0709, 0.0711, 0.0736, 0.0737, 0.0739, 0.0741, 0.0743, 0.0776, 0.0782, 0.0804, 0.0808, 0.0815, 0.0818, 0.0819, 0.0840, 0.0850, 0.0864, 0.0867, 0.0869, 0.0901, 0.0904, 0.0907, 0.0914, 0.0943, 0.0946, 0.1009, 0.1134).

\item[Second Dataset for Section \ref{sec:mle_eeiw}:] $\mathbf{x}$=(3.7, 3.11, 4.42, 3.28, 3.75, 2.96, 3.39, 3.31, 3.15, 2.81, 1.41, 2.76, 3.19, 1.59, 2.17, 3.51, 1.84, 1.61, 1.57, 1.89, 2.74, 3.27, 2.41, 3.09, 2.43, 2.53, 2.81, 3.31, 2.35, 2.77, 0.39, 2.79, 1.08, 2.88, 2.73, 2.85, 2.55, 2.17, 2.97, 3.68, 2.03, 2.82, 2.50, 1.47, 3.22, 2.83, 1.36, 1.84, 5.56, 1.12, 3.60, 3.11, 1.69, 4.90, 3.39, 1.59, 1.73, 1.71, 1.18, 4.38, 2.68, 4.91, 1.57, 2.00, 2.87, 3.19, 1.87, 2.95, 0.81, 1.22, 5.08, 1.69, 3.15, 2.97, 2.93, 3.33, 2.48, 1.25, 2.48, 2.03, 3.22, 2.55, 3.56, 2.38, 0.85, 1.80, 2.12, 3.65, 1.17, 2.17, 2.67, 4.20, 3.68, 4.70, 2.56, 2.59, 1.61, 2.05, 1.92, 0.98).

\item[Third Dataset for Section \ref{sec:mle_eeiw}:] $\mathbf{x}$=(17.88, 28, 92, 33, 41.52, 42.12, 45.60, 48.4, 51.84, 51.96, 54.12, 55.56, 67.8, 68.64, 68.88, 84.12, 93.12, 98.64, 105.12, 105.84, 127.92, 128.04, 173.4).

\end{description}

\printbibliography
\end{document}